\documentclass[letterpaper]{article} % DO NOT CHANGE THIS
\usepackage{aaai2027}  % DO NOT CHANGE THIS
\nocopyright
\usepackage[hyphens]{url}  % DO NOT CHANGE THIS
\usepackage{graphicx} % DO NOT CHANGE THIS
\usepackage{natbib}  % DO NOT CHANGE THIS AND DO NOT ADD ANY OPTIONS TO IT
\usepackage{caption} % DO NOT CHANGE THIS AND DO NOT ADD ANY OPTIONS TO IT
\usepackage{algorithm}
\usepackage{algorithmic}

\usepackage{newfloat}
\usepackage{listings}
\DeclareCaptionStyle{ruled}{labelfont=normalfont,labelsep=colon,strut=off} % DO NOT CHANGE THIS
\floatstyle{ruled}
\newfloat{listing}{tb}{lst}{}
\floatname{listing}{Listing}

\usepackage{booktabs}

\usepackage{algorithm}
\usepackage{amsmath}
\usepackage{amssymb}
\usepackage{mathtools}
\usepackage{pifont}
\usepackage{bm}
\usepackage{pdfpages}
\usepackage{microtype}
\usepackage{graphicx}
\usepackage{booktabs}  
\usepackage{multirow}
\usepackage{xcolor}     
\usepackage{newunicodechar}
\newunicodechar{❶}{\ding{172}}
\newunicodechar{❷}{\ding{173}}
\usepackage{colortbl}
\usepackage{bbding}
\usepackage{bm}
\definecolor{mygreen}{RGB}{46, 139, 87}  
\definecolor{myred}{RGB}{205, 92, 92}     
\definecolor{mygray}{gray}{0.6}  
\definecolor{lightyellow}{RGB}{255, 250, 205} 
\definecolor{algcomment}{RGB}{70,130,180}

\newcommand{\myconf}[1]{\textcolor{mygray}{\scriptsize #1}}
\newcommand{\up}[1]{\ensuremath{_{\textcolor{mygreen}{\uparrow #1}}}}
\newcommand{\down}[1]{\ensuremath{_{\textcolor{myred}{\downarrow #1}}}}
\newcommand{\second}[1]{\underline{#1}}
\newcommand{\best}[1]{\textbf{#1}}
\newtheorem{assumption}{Assumption}
\newtheorem{proposition}{Proposition}

\title{Rethinking One-Shot Federated Graph Learning: Training-Free Statistical Estimation}
\author{
    Shutong Zheng\textsuperscript{\rm 2},
    Sijia Chen\textsuperscript{\rm 1}\corresponding
}
\affiliations{
    \textsuperscript{\rm 1}The Hong Kong University of Science and Technology (Guangzhou)\\
    \textsuperscript{\rm 2}Sun Yat-sen University\\
    zhengsht29@mail2.sysu.edu.cn, sijiachen@hkust-gz.edu.cn
}
\begin{document}

\maketitle

\begin{abstract}
One-shot federated graph learning generally aims to train Graph Neural Networks (GNNs) across clients with disconnected subgraphs in a single communication round. Existing methods predominantly design advanced optimization strategies under the premise that local GNN training is indispensable. However, empirical observations reveal that under extreme non-IID conditions, local GNN training suffers from severe cross-client representation misalignment, becoming a major source of error rather than a remedy. Motivated by this, we reformulate one-shot FGL as a statistical estimation problem.
We propose SPEAR (Statistical Prototype Estimation with Adaptive Reliability), a completely training-free framework that directly computes topology-smoothed class prototypes from local graphs in the original feature space. The server then aggregates these prototypes using a sample-size-adaptive shrinkage estimator that down-weights unreliable local estimates, producing robust global class prototypes.
Extensive experiments across seven benchmarks demonstrate that SPEAR consistently achieves state-of-the-art accuracy under extreme heterogeneity. Moreover, SPEAR delivers at least an order-of-magnitude speedup over all baselines, reaching several orders of magnitude against generative and distillation-based methods.
Our findings suggest that training-free statistical estimation, rather than local GNN optimization, provides the key to robust and efficient one-shot federated graph learning. The code is available at \url{https://github.com/Yodeesy/SPEAR}.
\end{abstract}


\section{Introduction}
Federated Learning (FL) provides a privacy-preserving mechanism for distributed model training without exposing raw local data. While conventional FL typically assumes independent data samples, real-world applications, such as financial networks and biochemical interactions, often generate highly interconnected, graph-structured data. This practical necessity has driven the rapid development of Federated Graph Learning (FGL). Nevertheless, partitioning graph data across distinct clients inherently severs the global topology, causing severe structural heterogeneity that traditional FL architectures cannot effectively resolve.

\begin{figure}[t]
\centering 
\includegraphics[width=0.48\textwidth]{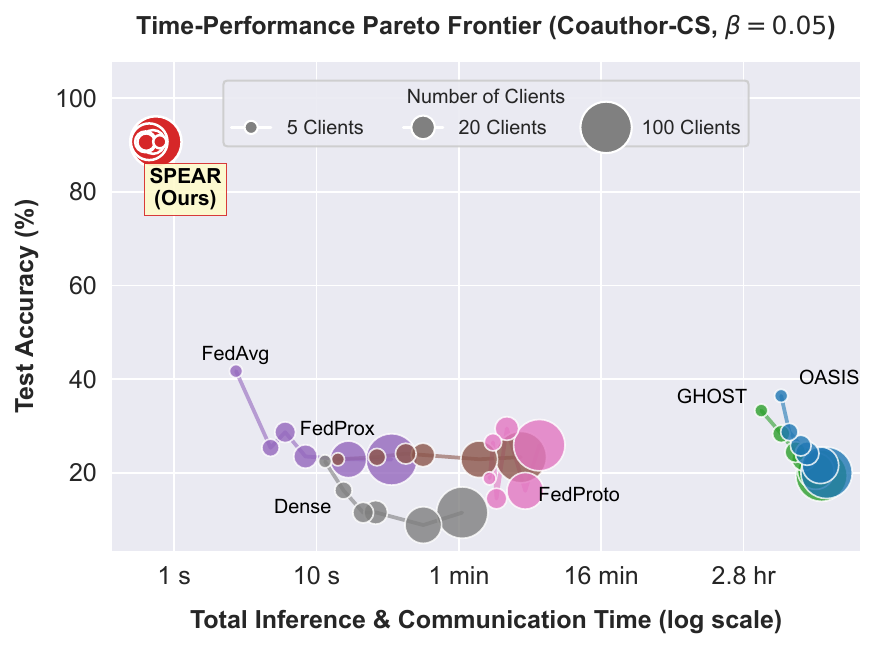}
\caption{
Time-Performance Pareto frontier on Coauthor-CS ($\beta = 0.05$). Bubble size indicates the number of clients. To ensure fair overhead comparison, traditional FL (e.g., FedAvg) runs 100 rounds, whereas one-shot methods complete in 1 round. SPEAR breaks the accuracy-efficiency trade-off, delivering superior performance and multi-magnitude speedups across all scales.
}
\label{fig:performance_and_time}
\end{figure}

To address this topological divergence, numerous advanced FGL frameworks have been proposed, such as FedSage~\cite{zhang2021subgraph}, FedTAD~\cite{zhu2024fedtad}, and FedATH~\cite{fu2025less}. Although effective at modeling heterogeneous graphs, these methods share a critical operational bottleneck: they require continuous, multi-round communication to iteratively align local models. This repetitive synchronization imposes prohibitive bandwidth and latency overheads, rendering them impractical for resource-constrained edge deployments. Consequently, one-shot FGL has emerged as a crucial alternative, strictly limiting the collaborative knowledge aggregation to a single communication round.

However, emerging studies (e.g., FedSD2C~\cite{zhang2024one}, GHOST~\cite{qian2025ghost}, and OASIS~\cite{wanoasis}) typically rely on training local GNNs to extract representations before performing a single server-side aggregation. In this work, we make a surprising empirical observation: under extreme non-IID conditions, local GNN training tends to overfit to client-specific label distributions and structural patterns, inducing severe cross-client representation misalignment. In the absence of iterative communication, these misaligned local representations cannot be effectively corrected, resulting in ineffective global aggregation and substantial performance degradation. This observation raises a fundamental question: is local parameter optimization truly necessary for extreme one-shot FGL?

To answer this question, we depart from the paradigm of local neural optimization and reformulate one-shot FGL from a purely statistical estimation perspective. We propose \textbf{SPEAR} (\textbf{S}tatistical \textbf{P}rototype \textbf{E}stimation with \textbf{A}daptive \textbf{R}eliability), a completely training-free framework for one-shot FGL under extreme non-IID. Locally, instead of training GNNs, SPEAR applies Personalized PageRank (PPR) as a graph low-pass filter to estimate topology-smoothed class prototypes directly in the original feature space. Globally, inspired by the James–Stein estimator, the server performs reliability-adaptive shrinkage estimation over these local prototypes, where the shrinkage strength is automatically determined by each client's local class sample size. This statistical estimator naturally suppresses unreliable estimates from data-scarce clients, yielding robust global prototypes without iterative communication. Finally, zero-gradient classification is achieved through direct cosine similarity between PPR-smoothed node features and the estimated global prototypes.

Overall, our contributions are summarized as follows:
\begin{itemize}
\item We present an empirical study showing that under extreme non-IID settings, existing training-based one-shot FGL methods consistently suffer from severe cross-client representation inconsistency, motivating a fundamental rethinking of local neural optimization in one-shot FGL.

\item We propose SPEAR, a completely training-free statistical inference framework for one-shot federated graph learning. SPEAR replaces iterative local GNN optimization with topology-smoothed prototype estimation and reliability-adaptive shrinkage estimation, enabling robust global prototype construction without gradient optimization or iterative communication.

\item Extensive experiments on seven benchmark datasets demonstrate that SPEAR consistently achieves state-of-the-art performance while exhibiting remarkable robustness under extreme heterogeneity, reducing computational overhead by orders of magnitude compared with existing training-based, distillation-based, and generative one-shot FGL methods (as illustrated in Fig.~\ref{fig:performance_and_time}).
\end{itemize}

\section{Related Work}

\subsection{Multi-Round Federated Graph Learning}
A major bottleneck in FGL is the iterative client-server communication used to mitigate the graph heterogeneity introduced when a global graph is partitioned across clients. Prior work reconstructs the resulting missing cross-client topology through repeated exchanges of, e.g., synthesized neighbors (FedSage~\cite{zhang2021subgraph}), topology-aware distillation (FedTAD~\cite{zhu2024fedtad}), or causally decoupled representations (FedATH~\cite{fu2025less}). Regardless of mechanism, each round requires retraining and re-transmission, and this cycle scales poorly with client count or graph size, becoming impractical under tight bandwidth or latency budgets.

\subsection{One-shot Federated Graph Learning}
One-shot FL compresses collaboration into a single round via distillation, generator training, or synthetic-data sharing~\cite{li2020practical, zhang2022dense, allouah2024revisiting, zeng2024one}, but extending this to graphs is nontrivial since node attributes are tightly coupled with topology. Recent one-shot FGL methods therefore adapt a GNN-centric pipeline to the graph domain: FedSD2C~\cite{zhang2024one} synthesizes condensed graphs to replace raw parameters, GHOST~\cite{qian2025ghost} maintains client-specific proxy models with topology-critical parameter consolidation, and OASIS~\cite{wanoasis} builds a topological codebook to generate synthetic graphs. Despite differing mechanisms, all still require optimizing a local GNN before any information leaves the client.

\subsection{Training-Free Statistical Learning}
A separate line of work questions whether training is necessary at all. In the centralized setting, Sato~\cite{sato2024trainingfree} shows a training-free GNN with labels as input features can match trained GCNs/GATs, suggesting propagation itself, not learned parameters, drives much of a GNN's power. This has reached federated learning via gradient-free aggregation: FedCGS~\cite{aaai2025fedcgs} builds a one-shot classifier from client image statistics, and Gaussian-Head OFL~\cite{iclr2026ghofl} extends this to text. Both, however, operate on non-graph data as independent feature vectors, discarding the relational structure central to graphs. This gap motivates SPEAR, to our knowledge, the first training-free statistical formulation of federated graph learning.

\section{Preliminaries and Motivation}
\begin{figure*}[htbp]
    \centering
    % === (a) Raw ===
    \begin{minipage}[b]{0.19\linewidth}
        \centering
        \includegraphics[width=\linewidth]{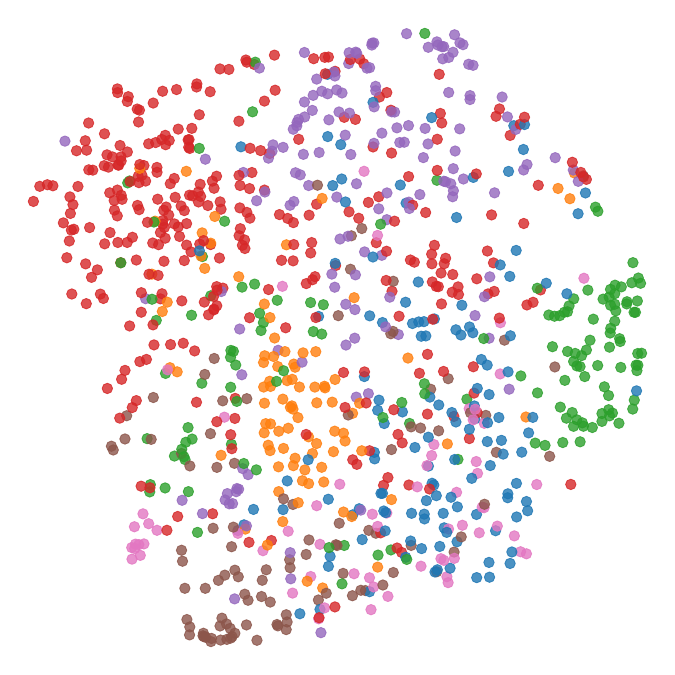}
        \centerline{\small (a) Raw Features}
    \end{minipage}
    \hfill
    % === (b) FedAvg ===
    \begin{minipage}[b]{0.19\linewidth}
        \centering
        \includegraphics[width=\linewidth]{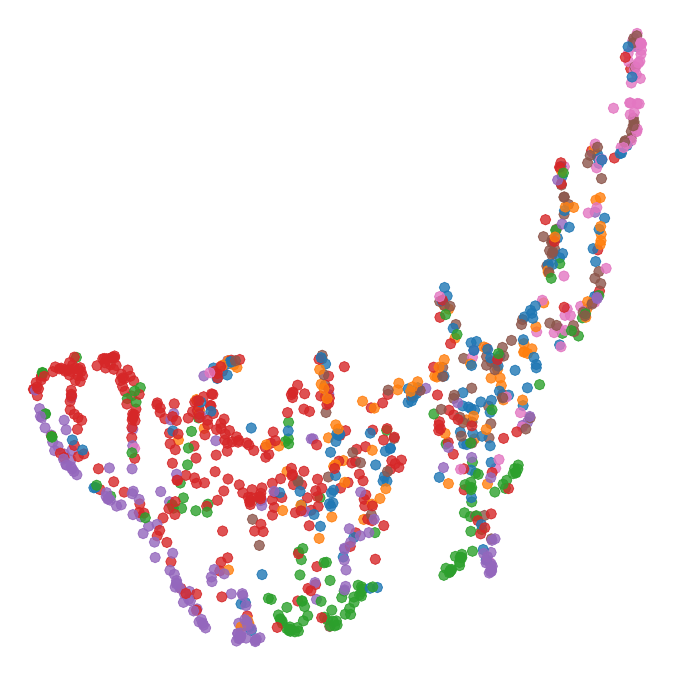}
        \centerline{\small (b) FedAvg}
    \end{minipage}
    \hfill
    % === (c) FedProto ===
    \begin{minipage}[b]{0.19\linewidth}
        \centering
        \includegraphics[width=\linewidth]{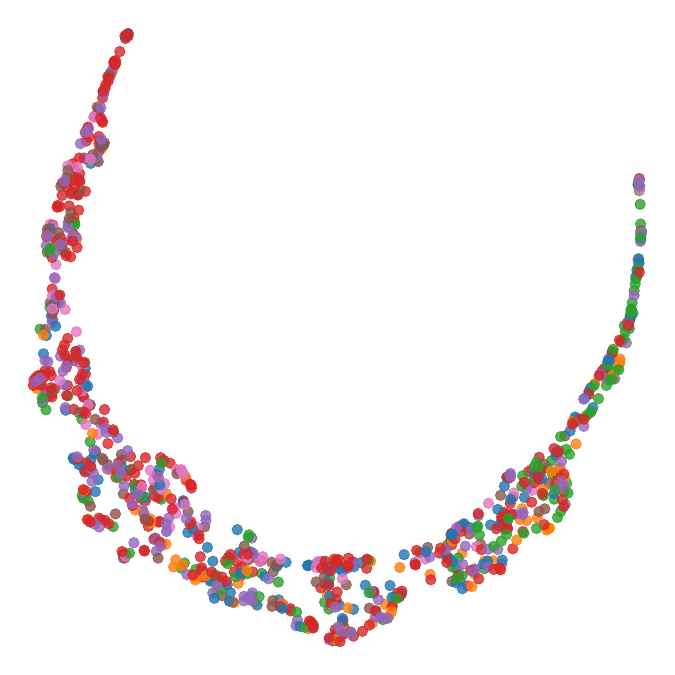}
        \centerline{\small (c) FedProto}
    \end{minipage}
    \hfill
    % === (d) FedGTA ===
    \begin{minipage}[b]{0.19\linewidth}
        \centering
        \includegraphics[width=\linewidth]{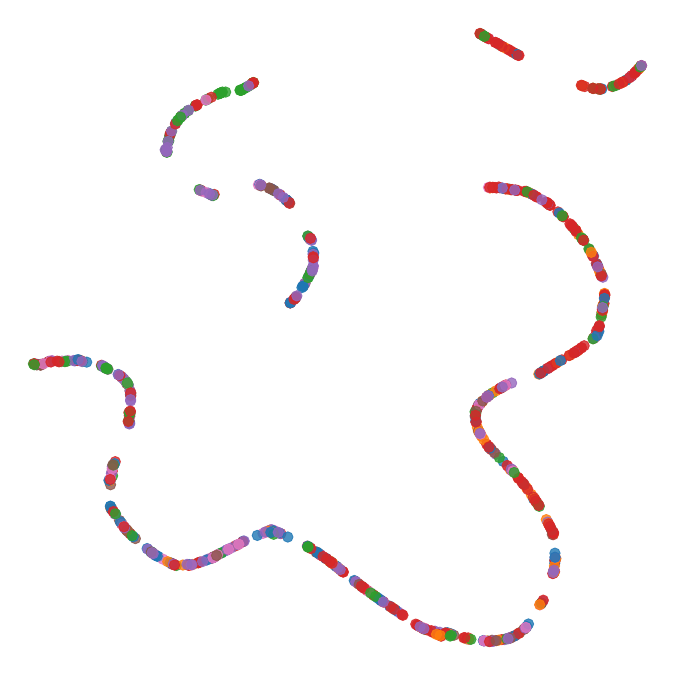}
        \centerline{\small (d) FedGTA}
    \end{minipage}
    \hfill
    % === (e) Silhouette Score ===
    \begin{minipage}[b]{0.22\linewidth}
        \centering
        \includegraphics[width=\linewidth]{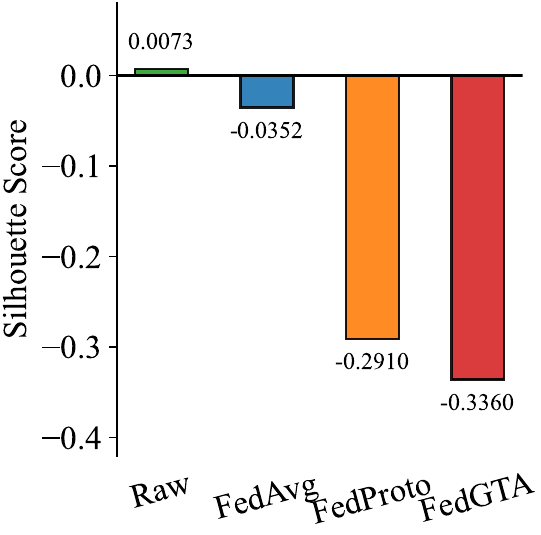}
        \centerline{\small (e) Silhouette Scores}
    \end{minipage}
    
    \caption{UMAP Visualization and quantitative evaluation of node representations under extreme non-IID conditions (Cora, $\beta=0.05$). Parameter-driven local optimization actively destroys the intrinsic structural alignment present in the raw features, leading to severe representation degradation.}
    \label{fig:motivation}
\end{figure*}

\subsection{Background}
We consider a federated system with $K$ clients, where each client $k$ owns a private local subgraph $\mathcal{G}_k = (\mathcal{V}_k, \mathcal{E}_k, \mathbf{X}_k, \mathbf{Y}_k)$. Here, $\mathcal{V}_k$ denotes the node set of size $N_k$, $\mathcal{E}_k$ is the edge set, $\mathbf{X}_k \in \mathbb{R}^{N_k \times F}$ is the node feature matrix with dimension $F$, and $\mathbf{Y}_k$ represents the labels. The topology is defined by an adjacency matrix $\mathbf{A}_k \in \{0,1\}^{N_k \times N_k}$. Under this distributed paradigm, cross-client edges are inherently unavailable. Furthermore, the label distribution $\mathcal{P}(\mathbf{Y}_k)$ follows a Dirichlet distribution parameterized by $\beta$, where a smaller $\beta$ induces increasingly severe structural and label heterogeneity (i.e., extreme non-IID conditions).

Standard FGL encodes local subgraphs via parameterized Graph Neural Networks (GNNs). In canonical forms (e.g., GCN or GraphSAGE), the $l$-th layer updates node representations $\mathbf{H}^{(l)}$ via:
\begin{equation}
\mathbf{H}^{(l+1)} = \sigma\left(\text{PROP}(\mathbf{A}_k) \mathbf{H}^{(l)} \mathbf{W}^{(l)}\right), \quad \mathbf{H}^{(0)} = \mathbf{X}_k
\end{equation}
where $\text{PROP}(\cdot)$ is a structural propagation operator (e.g., the normalized adjacency matrix), $\mathbf{W}^{(l)}$ is a trainable weight matrix, and $\sigma(\cdot)$ is a non-linear activation function.

In the \emph{one-shot} FGL setting, communication is restricted to a single round: clients transmit a compact summary of their local graphs exactly once, and the server must construct a global solution directly from these one-time summaries without subsequent refinement.

\subsection{Motivation}
\label{sec:motivation}
While locally optimizing GNN parameters before transmission is commonly assumed to extract robust features, we observe a striking phenomenon: under extreme non-IID conditions, local neural optimization is consistently associated with severe degradation in cross-client representation consistency. As visualized in the UMAP projections (Fig.~\ref{fig:motivation}), \emph{raw features} without any neural transformation naturally group into shared semantic regions regardless of client origins. However, local training substantially disrupts this intrinsic alignment. Even methods explicitly designed to handle heterogeneity, such as FedProto~\cite{tan2022fedproto} and FedGTA~\cite{li2023fedgta}, exhibit more severe cluster distortion than standard FedAvg~\cite{mcmahan2017communication}.

To quantitatively corroborate this, Fig.~\ref{fig:motivation} reports the Silhouette Score (cosine) for each representation space on Cora. Consistent with the visual evidence, raw features yield a modestly positive score, whereas gradient-based methods drive it heavily into the negative range (e.g., collapsing below $-0.29$ for FedProto and FedGTA). This confirms that the visual fragmentation reflects a genuine degradation of representation quality, rather than an artifact of UMAP's non-linear projection.

We observe consistent degradation trends across broader benchmarks and baselines (see Appendix E). This pervasive failure suggests that the bottleneck lies not in the specific design of local regularization, but in the reliance on local neural training itself. Motivated by this, we abandon local optimization entirely and propose SPEAR (Sec.~\ref{sec:method}), paving the way for a training-free statistical paradigm in FGL.

\section{Methodology}
\label{sec:method}
\begin{figure*}[t]
	\centering
	\includegraphics[width=\textwidth]{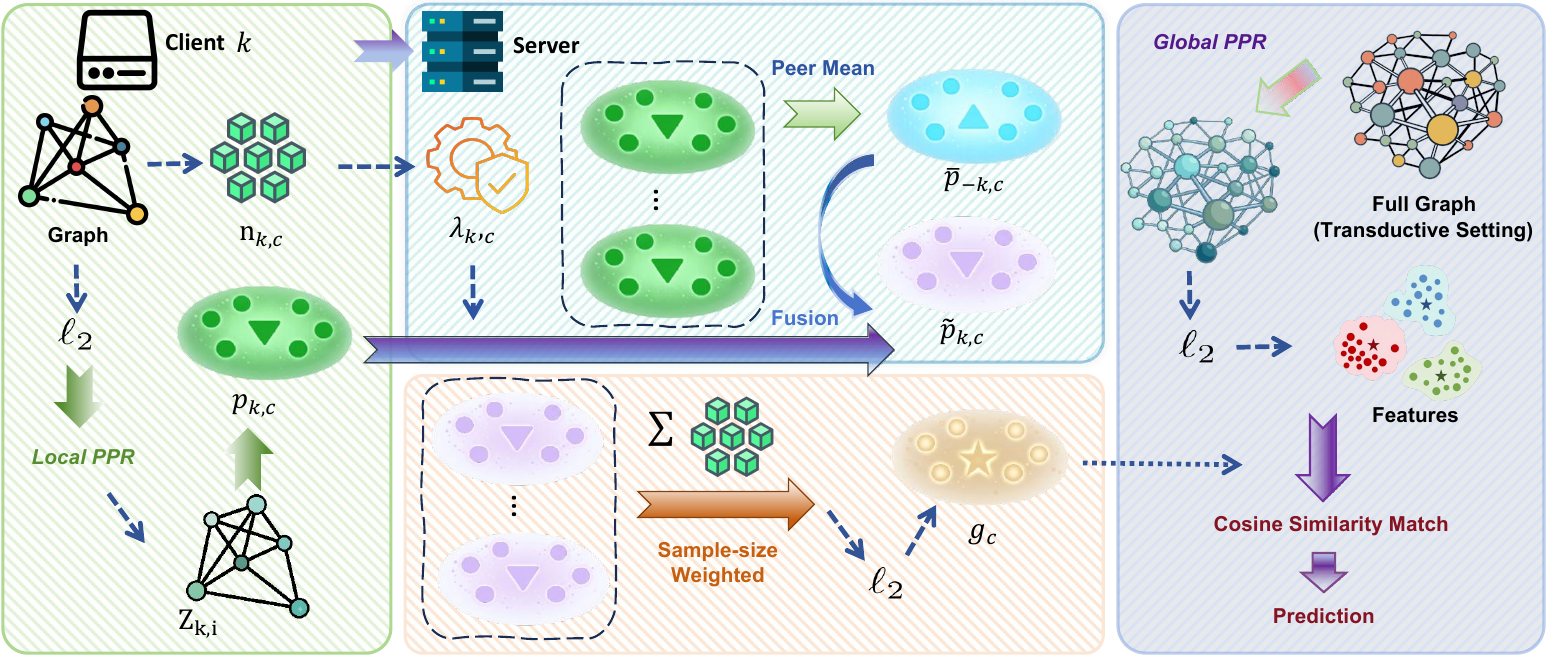}
    \caption{Architecture of SPEAR. (Left) Clients estimate local prototypes $\mathbf{p}_{k,c}$ via local topology smoothing (PPR) and transmit them alongside class sample sizes $n_{k,c}$. (Center) The server calibrates each local prototype via reliability-adaptive shrinkage (fusing it with the peer mean $\bar{\mathbf{p}}_{-k,c}$) and performs sample-size weighted aggregation to yield the global prototype $\mathbf{g}_c$. (Right) For inference, predictions are made via cosine similarity matching between $\mathbf{g}_c$ and the globally smoothed node features.}
    \label{fig:architecture}
\end{figure*}

\subsection{Topology-Smoothed Prototype Estimation}
\label{sec:clients}

Local parameter optimization is empirically associated with severe feature misalignment under extreme non-IID conditions (Sec.~\ref{sec:motivation}). Therefore, we retain only its topology propagation operator while discarding all trainable transformations and the non-linear activation $\sigma(\cdot)$ from the canonical GNN message-passing formulation. Instead, we rely exclusively on a training-free structural propagation operator to smooth the raw features.

Specifically, we employ a truncated Personalized PageRank (PPR) as our topological diffusion mechanism, following the personalized propagation scheme of APPNP~\cite{klicpera2019predict}. To ensure scale consistency across heterogeneous clients, client $k$ first applies $L_2$ normalization to its raw node features, yielding $\tilde{\mathbf{X}}_k$. We then compute the row-normalized transition matrix with self-loops as $\hat{\mathbf{A}}_k = \tilde{\mathbf{D}}^{-1}(\mathbf{A}_k + \mathbf{I})$, where $\tilde{\mathbf{D}}$ is the diagonal degree matrix of $\mathbf{A}_k + \mathbf{I}$.

The topology-smoothed feature matrix $\mathbf{Z}_k$ is then obtained through a finite $T$-step iterative propagation:
\begin{equation}
 \mathbf{Z}_k^{(t+1)} = (1 - \alpha) \tilde{\mathbf{X}}_k + \alpha \hat{\mathbf{A}}_k \mathbf{Z}_k^{(t)}, \quad t = 0, \dots, T-1 
\end{equation}
where we initialize $\mathbf{Z}_k^{(0)} = \tilde{\mathbf{X}}_k$. Here, $\alpha \in [0, 1)$ is the propagation weight assigned to the local neighborhood, with smaller $\alpha$ favoring the retention of the original features. In practice, we truncate the iteration to a small number of steps (e.g., $T=2$) as an efficient approximation. By propagating features through the local topology without any gradient-based updates, the final representation $\mathbf{Z}_k = \mathbf{Z}_k^{(T)}$ successfully captures structural semantics without introducing the parameter-driven drift.

Based on these smoothed representations, client $k$ constructs a local prototype for each observed class $c \in \mathcal{C}_k$ by averaging the representations of all locally available nodes belonging to that class:
\begin{equation}
 \mathbf{p}_{k,c} = \frac{1}{|\mathcal{V}_{k,c}|} \sum_{v_i \in \mathcal{V}_{k,c}} \mathbf{z}_{k,i} 
\end{equation}
where $\mathcal{V}_{k,c} = \{v_i \in \mathcal{V}_k \mid y_i = c\}$ denotes the set of nodes labeled as class $c$ in client $k$, and $\mathbf{z}_{k,i}$ is the $i$-th row of $\mathbf{Z}_k$.

Finally, instead of transmitting a bulky and misaligned GNN model, client $k$ only uploads its compact prototype set $\mathcal{P}_k = \{\mathbf{p}_{k,c} \mid c \in \mathcal{C}_k\}$ alongside the corresponding sample counts $n_{k,c} = |\mathcal{V}_{k,c}|$ to the central server. This strictly satisfies the communication-efficient one-shot constraint.

\begin{table*}[t]
\centering
\caption{Performance comparison on node classification tasks under the extreme non-IID setting (Dirichlet $\beta=0.05$ with $K=10$ clients). The abbreviations in the first column denote: \textbf{BL} (Baseline), and \textbf{OSFL} (One-shot Federated Learning). The best results are highlighted in \textbf{bold}, and the second-best results are \underline{underlined}. The small numbers indicate the performance gap compared to FedAvg (\textcolor{mygreen}{$\uparrow$} improvement, \textcolor{myred}{$\downarrow$} degradation).}
\label{tab:main_results}

\resizebox{\textwidth}{!}{%
\begin{tabular}{c|l||ccccccc}
\toprule
\textbf{Type} & \textbf{Methods} & \textbf{Cora} & \textbf{CiteSeer} & \textbf{PubMed} & \textbf{Amz-Comp} & \textbf{WikiCS} & \textbf{Coauthor-CS} & \textbf{ogbn-arxiv} \\ 
\midrule
% 基准 BL
BL & FedAvg \myconf{[AISTATS17]} & 34.74 & 36.08 & 61.92 & 37.70 & 19.47 & 25.33 & 14.58 \\ 
\midrule
% Traditional FL
\multirow{3}{*}{FL} 
& FedProx \myconf{[MLSys20]} & 28.24\down{6.50} & 30.48\down{5.60} & 54.62\down{7.30} & 24.74\down{12.96} & 4.22\down{15.25} & 23.33\down{2.00} & 13.37\down{1.21} \\
& FedOPT \myconf{[ICLR21]} & 29.68\down{5.06} & 21.82\down{14.26} & 43.86\down{18.06} & 12.42\down{25.28} & 10.57\down{8.90} & 7.76\down{17.57} & 2.14\down{12.44} \\
& MOON \myconf{[CVPR21]} & 27.98\down{6.76} & 22.12\down{13.96} & 45.50\down{16.42} & 23.71\down{13.99} & 4.13\down{15.34} & 11.44\down{13.89} & 11.57\down{3.01} \\ 
\midrule
% Traditional FGL
\multirow{4}{*}{FGL} 
& FedProto \myconf{[AAAI22]} & 30.50\down{4.24} & 24.56\down{11.52} & 52.86\down{9.06} & 37.66\down{0.04} & 6.67\down{12.80} & 26.51\up{1.18} & 5.20\down{9.38} \\
& FedPub \myconf{[ICML23]} & 21.16\down{13.58} & 19.16\down{16.92} & 41.16\down{20.76} & 37.70\down{0.00} & 17.49\down{1.98} & 8.51\down{16.82} & 10.16\down{4.42} \\
& FedGTA \myconf{[VLDB24]} & 27.06\down{7.68} & 20.92\down{15.16} & 42.98\down{18.94} & \second{38.22}\up{0.52} & 6.63\down{12.84} & 7.65\down{17.68} & 1.77\down{12.81} \\ 
& FedTAD \myconf{[IJCAI24]} & 22.92\down{11.82} & 22.48\down{13.60} & 46.96\down{14.96} & 31.71\down{5.99} & 16.25\down{3.22} & 4.62\down{20.71} & 1.27\down{13.31} \\
\midrule
% One-shot FL
\multirow{6}{*}{OSFL} 
& DENSE \myconf{[NeurIPS22]} & 25.10\down{9.64} & 18.36\down{17.72} & 41.70\down{20.22} & 14.94\down{22.76} & \second{17.83}\down{1.64} & 16.22\down{9.11} & 5.86\down{8.72} \\
& FedCVAE \myconf{[ICLR23]} & 14.78\down{19.96} & 17.17\down{18.91} & 33.68\down{28.24} & 10.86\down{26.84} & 17.36\down{2.11} & 11.97\down{13.36} & 11.52\down{3.06} \\
& FedSD2C \myconf{[NeurIPS24]} & 21.68\down{13.06} & 18.66\down{17.42} & 32.90\down{29.02} & 12.94\down{24.76} & 12.18\down{7.29} & 10.03\down{15.30} & 5.87\down{8.71} \\ 
& GHOST \myconf{[ICML25]} & 38.19\up{3.45} & 38.95\up{2.87} & 59.03\down{2.89} & 36.59\down{1.11} & 11.95\down{7.52} & 28.31\up{2.98} & \second{15.53}\up{0.95} \\
& OASIS \myconf{[NeurIPS25]} & \second{41.66}\up{6.92} & \second{41.29}\up{5.21} & \second{62.48}\up{0.56} & 36.83\down{0.87} & 12.46\down{7.01} & \second{28.70}\up{3.37} & OOM \\
\cmidrule{2-9}
% Ours 行高亮背景色
& \cellcolor{lightyellow}\textbf{SPEAR (Ours)} & \cellcolor{lightyellow}\best{63.35}\up{28.61} & \cellcolor{lightyellow}\best{65.47}\up{29.39} & \cellcolor{lightyellow}\best{74.06}\up{12.14} & \cellcolor{lightyellow}\best{76.42}\up{38.72} & \cellcolor{lightyellow}\best{63.76}\up{44.29} & \cellcolor{lightyellow}\best{90.81}\up{65.48} & \cellcolor{lightyellow}\best{37.24}\up{22.66} \\ 
\bottomrule
\end{tabular}%
}
\end{table*}

\subsection{Reliability-Adaptive Prototype Shrinkage}
\label{sec:server}

Upon receiving the local prototype sets $\mathcal{P}_k$ and the corresponding sample counts $n_{k,c}$ from all clients, the central server aims to construct a robust global prototype $\mathbf{g}_c$ for each class $c \in \mathcal{C}$. Under extreme non-IID conditions, local label distributions are highly skewed. Consequently, some local prototypes are computed from an abundant number of samples, while others are derived from only a handful of instances.

From a statistical estimation perspective, a local prototype $\mathbf{p}_{k,c}$ estimated from a scarce sample size $n_{k,c}$ exhibits high variance and significant epistemic uncertainty. A naive aggregation strategy is acutely vulnerable to these noisy, small-sample estimates, which can severely distort the global representation. To address this, we introduce a reliability-adaptive shrinkage mechanism. 

Specifically, to calibrate this noisy local estimate, we first construct a peer mean $\bar{\mathbf{p}}_{-k,c}$ by aggregating the prototypes of class $c$ from all other participating clients:
\begin{equation}
 \bar{\mathbf{p}}_{-k,c} = \frac{1}{|\mathcal{S}_c| - 1} \sum_{j \in \mathcal{S}_c \setminus \{k\}} \mathbf{p}_{j,c}
\end{equation}
where $\mathcal{S}_c = \{k \mid n_{k,c} > 0\}$. We adopt a simple unweighted average here, as more elaborate peer-selection strategies (e.g., similarity-based weighting) yield no consistent empirical improvement, as discussed in Appendix G. In the special case where only client $k$ observes class $c$ (i.e., $|\mathcal{S}_c| = 1$), we set $\bar{\mathbf{p}}_{-k,c} = \mathbf{p}_{k,c}$, effectively disabling shrinkage for this isolated observation. The calibrated (shrunk) prototype $\tilde{\mathbf{p}}_{k,c}$ is then derived by pulling the uncertain local observation toward the stable peer mean:
\begin{equation}\label{eq:shrinkage}
\tilde{\mathbf{p}}_{k,c} = (1 - \lambda_{k,c}) \mathbf{p}_{k,c} + \lambda_{k,c} \bar{\mathbf{p}}_{-k,c} 
\end{equation}
Here, $\lambda_{k,c} \in (0, 1)$ is a reliability-adaptive shrinkage factor. 

To formally justify this aggregation weighting, we formulate the prototype calibration as a statistical shrinkage estimation problem inspired by James-Stein estimation. Let $\mathbf{p}_c^*$ be the true, unobserved class centroid for class $c$ in the shared feature space. The local prototype extracted by client $k$, denoted as $\mathbf{p}_{k,c}$, is modeled as the true signal corrupted by local epistemic noise: $\mathbf{p}_{k,c} = \mathbf{p}_c^* + \varepsilon_{k,c}$.

We ground our analysis on two standard statistical assumptions:

\begin{assumption}[Unbiased Isotropic Noise]\label{ass:noise}
The local noise $\varepsilon_{k,c}$ is independent across clients with zero mean and finite isotropic variance, i.e., $\mathbb{E}[\varepsilon_{k,c}] = \mathbf{0}$ and $\mathrm{Cov}(\varepsilon_{k,c}) = \sigma_{k,c}^2 \mathbf{I}_F$.
\end{assumption}

\begin{assumption}[Variance Scaling]\label{ass:variance}
Following the Law of Large Numbers, the variance of the local estimate is strictly inversely proportional to the local sample size $n_{k,c}$, modeled as $\sigma_{k,c}^2 = \frac{\sigma_0^2}{n_{k,c}}$, where $\sigma_0^2$ is the base variance of a single node feature.
\end{assumption}

Given a peer reference anchor $\boldsymbol{\nu}_c$ that is unbiased ($\mathbb{E}[\boldsymbol{\nu}_c] = \mathbf{p}_c^*$) with variance $\mathrm{Cov}(\boldsymbol{\nu}_c) = \tau^2 \mathbf{I}_F$, we define the reliability-adaptive shrinkage estimator as $\tilde{\mathbf{p}}_{k,c} = (1 - \lambda_{k,c})\mathbf{p}_{k,c} + \lambda_{k,c}\boldsymbol{\nu}_c$. In our implementation, the theoretical anchor $\boldsymbol{\nu}_c$ is instantiated as the peer mean $\bar{\mathbf{p}}_{-k,c}$ defined above.

\begin{proposition}[Risk Bound of Shrinkage Estimation]\label{prop:risk}
Under Assumptions~\ref{ass:noise} and~\ref{ass:variance}, the expected Mean Squared Error (MSE) of the shrinkage estimator is minimized by the optimal scalar weight $\lambda_{k,c}^* = \frac{\sigma_{k,c}^2}{\sigma_{k,c}^2 + \tau^2}$. Furthermore, this optimal expected MSE is strictly less than the variance of the uncalibrated local estimate whenever $\tau^2 > 0$. The detailed proof is provided in Appendix C.
\end{proposition}

By substituting the variance model from Assumption~\ref{ass:variance} into the optimal weight derived in Proposition~\ref{prop:risk}, we obtain:
$$ \lambda_{k,c}^* = \frac{\sigma_0^2 / n_{k,c}}{\sigma_0^2 / n_{k,c} + \tau^2} = \frac{1}{1 + (\tau^2/\sigma_0^2) n_{k,c}} $$
Letting $\gamma \triangleq \frac{\tau^2}{\sigma_0^2}$, we exactly recover the formulation used in our mechanism:
\begin{equation}\label{eq:lambda}
    \lambda_{k,c} = \frac{1}{1 + \gamma n_{k,c}}
\end{equation}
where $\gamma > 0$ is a scaling hyperparameter naturally representing the variance ratio between the peer anchor and the local base noise. A smaller sample size $n_{k,c}$ yields a larger $\lambda_{k,c}$, mathematically indicating higher epistemic uncertainty and enforcing a stronger shrinkage toward the peer consensus.

Through this formulation, prototypes with abundant samples ($\lambda_{k,c} \to 0$) dominate their own representations, whereas highly unreliable estimates ($\lambda_{k,c} \to 1$) are strongly regularized by the peer consensus. 

Finally, the server aggregates these calibrated prototypes via a sample-weighted average and applies $L_2$ normalization to construct the final global prototype $\mathbf{g}_c$:
\begin{equation}\label{eq:global_agg}
 \mathbf{g}_c = \text{Norm}\left( \frac{1}{\sum_{k \in \mathcal{S}_c} n_{k,c}} \sum_{k \in \mathcal{S}_c} n_{k,c} \tilde{\mathbf{p}}_{k,c} \right) 
\end{equation}
where $\text{Norm}(\mathbf{x}) = \mathbf{x} / \|\mathbf{x}\|_2$. While inverse-variance weighting ($\lambda_{k,c}=0$) is optimal under idealized i.i.d. noise, real-world truncated subgraphs induce finite-sample structural bias; combining local shrinkage (Eq.~\ref{eq:shrinkage}) with sample-weighted aggregation (Eq.~\ref{eq:global_agg}) jointly regularizes extreme client outliers and weights reliable observations.

\textbf{Global Inference.} 
During the evaluation phase, the central server utilizes the constructed global prototype set $\mathcal{G} = \{\mathbf{g}_c\}_{c=1}^{C}$ as a non-parametric classifier. For a target node $v_i$ with its topology-smoothed feature $\mathbf{z}_i$ (extracted via the same PPR diffusion scheme defined in Sec.~\ref{sec:clients}, applied over the test graph following standard transductive FGL benchmarks, or locally on each $\mathcal{G}_k$ under strict decentralized testing), the prediction $\hat{y}_i$ is generated by maximizing the cosine similarity between the $L_2$-normalized node representation and the global prototypes:
\begin{equation}
 \hat{y}_i = \arg\max_{c \in \mathcal{C}} \left( \frac{\mathbf{z}_i}{\|\mathbf{z}_i\|_2} \cdot \mathbf{g}_c \right) 
\end{equation}
This direct similarity matching completely bypasses parameterized layers, finalizing the unified training-free paradigm of SPEAR.

\textbf{Complexity Analysis.} Unlike training-based FGL requiring iterative backpropagation, SPEAR is entirely gradient-free. Client-side $T$-step PPR and prototype estimation take $\mathcal{O}(T |\mathcal{E}_k| F + N_k F)$ computations. Communication strictly involves a one-shot transmission of $\mathcal{O}(C F)$ per client. Server-side shrinkage requires only $\mathcal{O}(K C F)$ operations. This lightweight design directly underlies the substantial efficiency gains validated in Sec.~\ref{sec:performance} (Details in Appendix B).

\section{Experiments}
\label{sec:experiments}

\begin{table*}[t]
\centering
\caption{Robustness comparison under edge perturbation ($\rho=0.5$) and feature perturbation ($\eta=0.5$) on four representative datasets. Subscripts denote the absolute performance deviation (\up{} improvement, \down{} degradation) compared to the clean setting.}
\label{tab:perturbation}
\resizebox{\textwidth}{!}{%
\begin{tabular}{l || cc | cc | cc | cc }
\toprule
\multirow{2}{*}{\textbf{Methods}} & \multicolumn{2}{c|}{\textbf{Cora}} & \multicolumn{2}{c|}{\textbf{CiteSeer}} & \multicolumn{2}{c|}{\textbf{PubMed}} & \multicolumn{2}{c}{\textbf{ogbn-arxiv}} \\
\cmidrule{2-9}
& \textbf{Edge ($\rho=0.5$)} & \textbf{Feat. ($\eta=0.5$)} & \textbf{Edge ($\rho=0.5$)} & \textbf{Feat. ($\eta=0.5$)} & \textbf{Edge ($\rho=0.5$)} & \textbf{Feat. ($\eta=0.5$)} & \textbf{Edge ($\rho=0.5$)} & \textbf{Feat. ($\eta=0.5$)} \\
\midrule
FedAvg  & 31.90\down{2.84} & 31.90\down{2.84} & 18.10\down{17.98}& 18.20\down{17.88}& 42.40\down{19.52}& 48.30\down{13.62}& 22.28\up{7.70}  & 10.22\down{4.36} \\
DENSE   & 16.40\down{8.70} & 14.90\down{10.20}& 18.20\down{0.16} & 18.20\down{0.16} & 41.60\down{0.10} & 40.70\down{1.00} & 3.48\down{2.38} & 3.78\down{2.08} \\
FedCVAE & 15.10\up{0.32}   & 31.90\up{17.12}  & 17.20\up{0.03}   & 13.80\down{3.37} & 41.00\up{7.32}   & 40.80\up{7.12}   & 21.56\up{10.04} & 12.55\up{1.03} \\
FedSD2C & 14.90\down{6.78} & 31.85\up{10.17}  & 16.90\down{1.76} & 16.84\down{1.82} & 41.20\up{8.30}   & 40.70\up{7.80}   & 7.87\up{2.00}   & 18.87\up{13.00} \\
GHOST   & 32.75\down{5.44} & 33.46\down{4.73} & 40.60\up{1.65}   & 36.85\down{2.10} & 48.94\down{10.09}& 47.11\down{11.92}& 17.54\up{2.01}  & 12.79\down{2.74} \\
OASIS   & 47.57\up{5.91}   & 41.34\down{0.32} & 46.14\up{4.85}   & 39.10\down{2.19} & 50.21\down{12.27}& 49.25\down{13.23}& OOM             & OOM \\
\midrule
\rowcolor{lightyellow}
\textbf{SPEAR} & \textbf{62.73}\down{0.62} & \textbf{60.53}\down{2.82} & \textbf{65.31}\down{0.16} & \textbf{61.53}\down{3.94} & \textbf{73.67}\down{0.39} & \textbf{74.41}\up{0.35} & \textbf{37.23}\down{0.01} & \textbf{36.37}\down{0.87} \\
\bottomrule
\end{tabular}%
}
\end{table*}

\subsection{Experimental Setup}
We evaluate our framework on diverse node classification benchmarks to demonstrate its superior performance. Implementation details can be found in Appendix D.

\textbf{Datasets.} To ensure a thorough evaluation, we employed seven representative graph datasets that cover diverse domains and scales, including Cora~\citep{mccallum2000automating}, CiteSeer~\citep{giles1998citeseer}, PubMed~\citep{sen2008collective}, Amazon-Computers~\citep{shchur2018pitfalls}, Wiki-CS~\citep{mernyei2020wikics}, Coauthor-CS~\citep{shchur2018pitfalls}, and ogbn-arxiv~\citep{hu2020open}. Detailed dataset information and splits for these datasets are provided in Appendix D.

\textbf{Baselines.}
We compare our method with four traditional FL methods:
(1) \textbf{FedAvg}~\myconf{[AISTATS17]}~\citep{mcmahan2017communication}, 
(2) \textbf{FedProx}~\myconf{[MLSys20]}~\citep{li2020fedprox}, 
(3) \textbf{FedOPT}~\myconf{[ICLR21]}~\citep{reddi2021adaptive},
(4) \textbf{MOON}~\myconf{[CVPR21]}~\citep{li2021moon};
four popular FGL methods: 
(5) \textbf{FedProto}~\myconf{[AAAI22]}~\citep{tan2022fedproto}, 
(6) \textbf{FedPub}~\myconf{[ICML23]}~\citep{baek2023personalized}, 
(7) \textbf{FedGTA}~\myconf{[VLDB24]}~\citep{li2023fedgta},
(8) \textbf{FedTAD}~\myconf{[IJCAI24]}~\citep{zhu2024fedtad};
five One-Shot FGL methods: 
(9) \textbf{DENSE}~\myconf{[NeurIPS22]}~\citep{zhang2022dense}, 
(10) \textbf{FedCVAE}~\myconf{[ICLR23]}~\citep{heinbaugh2023data}, 
(11) \textbf{FedSD2C}~\myconf{[NeurIPS24]}~\citep{zhang2024one}, 
(12) \textbf{GHOST}~\myconf{[ICML25]}~\citep{qian2025ghost},
(13) \textbf{OASIS}~\myconf{[NeurIPS25]}~\citep{wanoasis}.

For fair comparison, all methods (including SPEAR) are evaluated under the identical standard transductive protocol: after training or aggregation, the server-side global model is evaluated centrally on the complete test graph.

\subsection{Performance Comparison}
\label{sec:performance}
\textbf{General Classification Performance.} 
Tab.~\ref{tab:main_results} reports the node classification accuracy. Under extreme non-IID settings ($\beta=0.05$), gradient-based methods degrade severely, consistent with the representation misalignment identified in Sec.~\ref{sec:motivation}. In contrast, SPEAR consistently outperforms all multi-round and one-shot baselines by a significant margin. Crucially, these gains stem not from architectural complexity, but from abandoning local optimization entirely. This supports our hypothesis that local training itself becomes the dominant source of error, a claim further validated in Sec.~\ref{sec:ablation}.

\textbf{Efficiency and Communication Overhead.}
\begin{table}[t]
\centering
\caption{Efficiency comparison on four representative datasets. Total federated training time (top) and communication cost (bottom) are reported for each method. The multiplier indicates SPEAR's speedup relative to FedAvg.}
\label{tab:efficiency}
\resizebox{\columnwidth}{!}{%
\begin{tabular}{l || cccc }
\toprule
\textbf{Methods} & \textbf{Cora} & \textbf{CiteSeer} & \textbf{PubMed} & \textbf{ogbn-arxiv} \\
\midrule
\multirow{2}{*}{FedAvg} & 66s & 68s & 67s & 80s \\
                        & 2814MB & 7246MB & 984MB & 330MB \\
\midrule
\multirow{2}{*}{GHOST}  & 71s & 102s & 5716s & 20028s \\
                        & 16.0MB & 38.0MB & 7.0MB & 3.0MB \\
\midrule
\multirow{2}{*}{OASIS}  & 179s & 189s & 7004s & OOM \\
                        & 44.0MB & 111MB & 17.0MB & - \\
\midrule
\rowcolor{lightyellow}
\textbf{SPEAR}  & \textbf{0.86s} (\textcolor{teal}{$\mathbf{77\times}$}) & \textbf{0.61s} (\textcolor{teal}{$\mathbf{111\times}$}) & \textbf{0.83s} (\textcolor{teal}{$\mathbf{81\times}$}) & \textbf{1.92s} (\textcolor{teal}{$\mathbf{42\times}$}) \\
\rowcolor{lightyellow}
\textbf{(Ours)} & \textbf{7.3KB} & \textbf{18.8KB} & \textbf{2.0KB} & \textbf{7.5KB} \\
\bottomrule
\end{tabular}%
}
\end{table}

As reported in Tab.~\ref{tab:efficiency}, SPEAR achieves orders-of-magnitude acceleration and bandwidth reduction. By transmitting only topology-smoothed prototypes instead of model parameters or gradients, communication drops from megabytes to kilobytes (e.g., 7246 MB $\rightarrow$ 18.8 KB on CiteSeer). Furthermore, eliminating local backpropagation and multi-round synchronization reduces wall-clock time by up to $111\times$ compared to multi-round FedAvg, outpacing even the fastest one-shot baseline (GHOST). This establishes SPEAR as a highly practical solution for resource-constrained edge environments.

\begin{table*}[t]
\centering
\caption{Ablation study of SPEAR across seven datasets. The full model is systematically degraded by removing specific components to assess their individual contributions. Subscripts denote the absolute performance degradation (\down{}) relative to the complete \textit{SPEAR} architecture.}
\label{tab:ablation}
\resizebox{\textwidth}{!}{%
\begin{tabular}{l | ccccccc }
\toprule
\textbf{Variants} & \textbf{Cora} & \textbf{CiteSeer} & \textbf{PubMed} & \textbf{Amz-Comp} & \textbf{WikiCS} & \textbf{Coauthor-CS} & \textbf{ogbn-arxiv} \\
\midrule
\rowcolor{lightyellow}
\textbf{SPEAR (Full Model)}     & \textbf{63.35} & \textbf{65.47} & \textbf{74.06} & \textbf{76.42} & \textbf{63.76} & \textbf{90.81} & \textbf{37.24} \\
\midrule
\textbf{w/o PPR}                & 59.00\down{4.35} & 62.03\down{3.44} & 72.60\down{1.46} & 74.32\down{2.10} & 61.01\down{2.75} & 89.85\down{0.96} & 33.78\down{3.46} \\
\textbf{w/o Shrinkage}          & 55.13\down{8.22} & 59.27\down{6.20} & 65.20\down{8.86} & 62.84\down{13.58}& 60.16\down{3.60} & 87.14\down{3.67} & 32.96\down{4.28} \\
\textbf{w/o Both (Baseline)}    & 59.70\down{3.65} & 61.70\down{3.77} & 72.30\down{1.76} & 74.33\down{2.09} & 60.99\down{2.77} & 89.84\down{0.97} & 33.79\down{3.45} \\
\bottomrule
\end{tabular}%
}
\end{table*}

\subsection{Robustness Analysis}
\label{sec:robustness}
Unless otherwise specified, all experiments in this section follow the same setup as the main results (Sec.~\ref{sec:performance}). Full results are provided in Appendix F.

\begin{figure}[t]
	\centering
	% === 左图 ===
	\begin{minipage}[b]{0.48\linewidth}
		\centering
		\includegraphics[width=\linewidth]{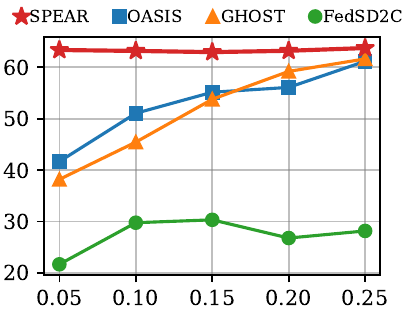}
		\centerline{\small (a) Varying $\beta$ in Dir($\beta$)}
	\end{minipage}
	\hfill % 左右推开
	% === 右图 ===
	\begin{minipage}[b]{0.48\linewidth}
		\centering
		\includegraphics[width=\linewidth]{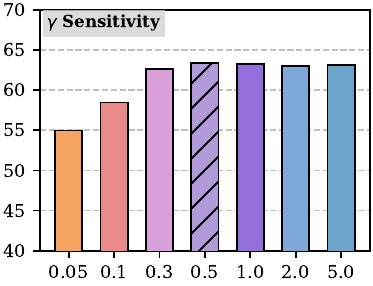}
		\centerline{\small (b) $\gamma$ Sensitivity}
	\end{minipage}
    \caption{Robustness analysis of SPEAR on Cora.}
	\label{fig:robustness}
\end{figure}

\textbf{Robustness to Structural and Feature Perturbations.}
As shown in Tab.~\ref{tab:perturbation}, SPEAR remains highly stable under both structural noise (mixed edge addition/deletion, $\rho=0.5$) and feature noise (random zero-masking, $\eta=0.5$), with degradation within 0.62\% and 3.94\%, respectively, across all four datasets. This robustness stems from PPR's role as a training-free topological low-pass filter: unlike gradient-based GNNs, which risk overfitting to erroneous edges during training, PPR requires no optimization and compensates for missing information through neighborhood aggregation.

\textbf{Hyperparameter Stability.} 
SPEAR is largely insensitive to its hyperparameters. As shown in Fig.~\ref{fig:robustness}(b), accuracy remains stable across an order of magnitude of $\gamma$, only degrading as $\gamma \to 0$ (which enforces excessive shrinkage $\lambda_{k,c} \to 1$ toward the unweighted peer mean). While $T$ exhibits similarly stable trends across all datasets, in contrast, $\alpha$ reveals a graph-dependent trade-off: increasing $\alpha$ can further improve accuracy on homophilous datasets (e.g., Cora), but degrades performance on heterophilous graphs (e.g., Texas, Actor), as stronger neighborhood aggregation amplifies noise from dissimilar neighbors. We therefore adopt a conservative, dataset-agnostic $\alpha$ throughout our experiments to prioritize robustness across diverse graph structures.

\textbf{Robustness to Data Heterogeneity and Scalability.} 
Beyond the extreme setting ($\beta=0.05$) in the main results, Fig.~\ref{fig:robustness}(a) sweeps $\beta$ from 0.05 to 0.25. SPEAR maintains a performance advantage over all baselines across this range, with the margin most pronounced under extreme heterogeneity and narrowing as $\beta$ increases. Its one-shot, training-free design further scales seamlessly with the number of clients, avoiding the synchronization bottlenecks of multi-round optimization.

Appendix H further examines SPEAR's relative advantage across a range of heterogeneity levels (varying Dirichlet $\beta$), its behavior on heterophilous graphs, and the limitations of the proposed training-free statistical paradigm.

\subsection{Ablation Study}
\label{sec:ablation}
To isolate the source of SPEAR's gains, we systematically ablate its components in Tab.~\ref{tab:ablation}. Notably, the \textit{Baseline} variant (\textit{w/o Both})—which simply computes global class centroids in the raw feature space without PPR or shrinkage—already substantially outperforms all training-based baselines in Tab.~\ref{tab:main_results}, confirming that bypassing local GNN optimization is the primary driver of robustness under extreme skew.

Comparing the remaining variants reveals a synergistic interaction. Removing reliability-adaptive calibration altogether (\textit{w/o Shrinkage}, which replaces Eqs.~(\ref{eq:shrinkage})--(\ref{eq:global_agg}) with unweighted client averaging $\frac{1}{|\mathcal{S}_c|}\sum_{k \in \mathcal{S}_c}\mathbf{p}_{k,c}$) severely degrades performance below the baseline (e.g., $74.33\% \rightarrow 62.84\%$ on Amazon-Computers), demonstrating that smoothed local prototypes are vulnerable to small-sample noise clients unless properly calibrated by sample-size reliability. Conversely, shrinkage alone (\textit{w/o PPR}) yields negligible improvements well within random variance. This implies shrinkage is not a generic averaging mechanism, but relies on informative, topology-smoothed prototypes to function effectively.

Thus, the two modules are complementary. PPR exploits local topology to improve the quality of prototype estimation, while shrinkage reduces the estimation uncertainty caused by highly heterogeneous local observations. Together, they refine the training-free estimator, enabling SPEAR to consistently achieve more reliable global prototypes without sacrificing its computational efficiency.

\subsection{Representation Analysis}
\label{sec:representation_analysis}

\begin{figure}[t]
	\centering
	% === 左图 ===
	\begin{minipage}[b]{0.48\linewidth}
		\centering
		\includegraphics[width=\linewidth]{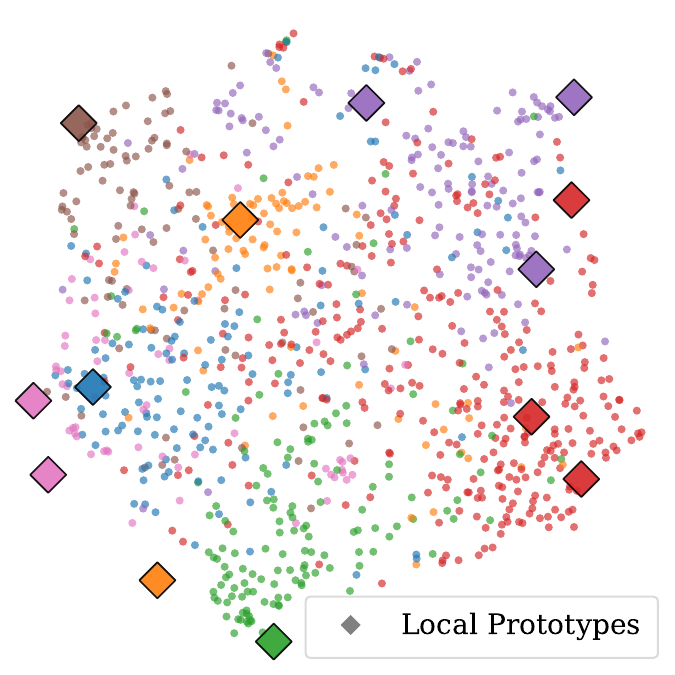}
		\centerline{\small (a) Local}
	\end{minipage}
	\hfill % 左右推开
	% === 右图 ===
	\begin{minipage}[b]{0.48\linewidth}
		\centering
		\includegraphics[width=\linewidth]{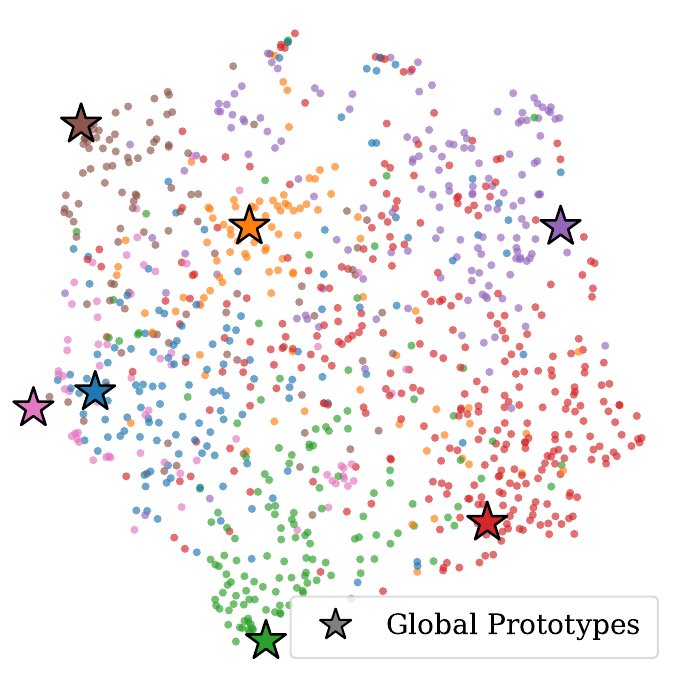}
		\centerline{\small (b) Global}
	\end{minipage}
    \caption{UMAP of SPEAR representations on Cora ($\beta=0.05$). Background scatter points are PPR-smoothed features. (a) Diamonds: uncalibrated local prototypes. (b) Stars: global prototypes calibrated via reliability-adaptive shrinkage.}
	\label{fig:representation_of_spear_cora}
\end{figure}

Unlike the fragmentation caused by parameter-driven methods (Fig.~\ref{fig:motivation}), Fig.~\ref{fig:representation_of_spear_cora} shows that SPEAR maintains cohesive representations. Fig.~\ref{fig:representation_of_spear_cora}(a) reveals that local prototypes (diamonds) initially exhibit dispersion due to data skewness. However, after aggregation (Fig.~\ref{fig:representation_of_spear_cora}(b)), they are calibrated into robust global prototypes (stars) tightly aligned with the true class regions. Quantitatively, the average cosine distance to the true centroids drops from $0.3217$ (local) to $0.1997$ (global), confirming that our shrinkage mechanism effectively corrects spatial bias and suppresses noise.

\section{Conclusion}

This work challenges the prevailing assumption that local GNN training is strictly necessary for one-shot federated graph learning: we find that under extreme non-IID conditions, parameter-driven local optimization is consistently associated with severe cross-client representation misalignment rather than a remedy for it. Guided by this finding, we introduce SPEAR (\textbf{S}tatistical \textbf{P}rototype \textbf{E}stimation with \textbf{A}daptive \textbf{R}eliability), a lightweight, training-free framework that reformulates one-shot FGL as a statistical estimation problem, extracting topology-smoothed local prototypes and calibrating them globally via reliability-adaptive shrinkage. Extensive evaluations across seven benchmarks confirm that SPEAR achieves superior classification accuracy under extreme heterogeneity while reducing computational and communication overhead by orders of magnitude. Our results suggest that lightweight statistical estimation offers a robust, communication-efficient alternative to conventional neural optimization for highly skewed distributed graphs. We hope SPEAR provides a reliable and transparent baseline for future explorations of training-free federated graph learning.

\bigskip

\bibliography{aaai2027}

@inproceedings{tan2022fedproto,
  title={{FedProto}: Federated Prototype Learning across Heterogeneous Clients},
  author={Tan, Yue and Long, Guodong and Liu, Lu and Zhou, Tianyi and Lu, Qinghua and Jiang, Jing and Zhang, Chengqi},
  booktitle={Proceedings of the AAAI Conference on Artificial Intelligence},
  volume={36},
  number={8},
  pages={8432--8440},
  year={2022}
}

@inproceedings{mcmahan2017communication,
	title={Communication-efficient learning of deep networks from decentralized data},
	author={McMahan, Brendan and Moore, Eider and Ramage, Daniel and Hampson, Seth and y Arcas, Blaise Aguera},
	booktitle={Proceedings of the International Conference on Artificial intelligence and statistics},
	pages={1273--1282},
	year={2017},
	//organization={PMLR}
}

@inproceedings{li2020fedprox,
	title={Federated optimization in heterogeneous networks},
	author={Li, Tian and Sahu, Anit Kumar and Zaheer, Manzil and Sanjabi, Maziar and Talwalkar, Ameet and Smith, Virginia},
	booktitle={Proceedings of the Machine learning and Systems Conference},
	volume={2},
	pages={429--450},
	year={2020}
}

@inproceedings{li2021moon,
	author       = {Qinbin Li and
	Bingsheng He and
	Dawn Song},
	title        = {Model-Contrastive Federated Learning},
	booktitle    = {Proceedings of {IEEE}/CVF Conference on Computer Vision and Pattern Recognition},
	pages        = {10713--10722},
	//publisher    = {Computer Vision Foundation / {IEEE}},
	year         = {2021},
	//url          = {https://openaccess.thecvf.com/content/CVPR2021/html/Li\_Model-Contrastive\_Federated\_Learning\_CVPR\_2021\_paper.html},
	//doi          = {10.1109/CVPR46437.2021.01057},
	//timestamp    = {Mon, 18 Jul 2022 16:47:41 +0200},
	//biburl       = {https://dblp.org/rec/conf/cvpr/LiHS21.bib},
	//bibsource    = {dblp computer science bibliography, https://dblp.org}
}

@inproceedings{
fu2025less,
title={Less is More: Federated Graph Learning with Alleviating Topology Heterogeneity from A Causal Perspective},
author={Lele Fu and Bowen Deng and Sheng Huang and Tianchi Liao and Shirui Pan and Chuan Chen},
booktitle={Proceedings of the International Conference on Machine Learning},
year={2025}
}

@article{zhang2021subgraph,
  title={Subgraph federated learning with missing neighbor generation},
  author={Zhang, Ke and Yang, Carl and Li, Xiaoxiao and Sun, Lichao and Yiu, Siu Ming},
  journal={Proceedings of the Advances in Neural Information Processing Systems},
  pages={6671--6682},
  year={2021}
}

@inproceedings{zhu2024fedtad,
  title={FedTAD: Topology-aware Data-free Knowledge Distillation for Subgraph Federated Learning},
  author={Zhu, Yinlin and Li, Xunkai and Wu, Zhengyu and Wu, Di and Hu, Miao and Li, Rong-Hua},
  booktitle={Proceedings of the International Joint Conference on Artificial Intelligence},
  pages={5716-5724},
  year={2024}
}

@inproceedings{baek2023personalized,
  title={Personalized subgraph federated learning},
  author={Baek, Jinheon and Jeong, Wonyong and Jin, Jiongdao and Yoon, Jaehong and Hwang, Sung Ju},
  booktitle={International conference on machine learning},
  pages={1396--1415},
  year={2023},
  organization={PMLR}
}

@article{iclr2026ghofl,
  title={The Gaussian-Head OFL Family: One-Shot Federated Learning from Client Global Statistics},
  author={Turazza, Fabio and Picone, Marco and Mamei, Marco},
  journal={arXiv preprint arXiv:2602.01186},
  year={2026}
}

@article{
sato2024trainingfree,
title={Training-free Graph Neural Networks and the Power of Labels as Features},
author={Ryoma Sato},
journal={Transactions on Machine Learning Research},
issn={2835-8856},
year={2024},
url={https://openreview.net/forum?id=7DzU88VrNU},
note={}
}

@inproceedings{aaai2025fedcgs,
  title={Capture global feature statistics for one-shot federated learning},
  author={Guan, Zenghao and Zhou, Yucan and Gu, Xiaoyan},
  booktitle={Proceedings of the AAAI Conference on Artificial Intelligence},
  volume={39},
  number={16},
  pages={16942--16950},
  year={2025}
}

@article{allouah2024revisiting,
  title={Revisiting ensembling in one-shot federated learning},
  author={Allouah, Youssef and Dhasade, Akash and Guerraoui, Rachid and Gupta, Nirupam and Kermarrec, Anne-Marie and Pinot, Rafael and Pires, Rafael and Sharma, Rishi},
  journal={Advances in Neural Information Processing Systems},
  volume={37},
  pages={68500--68527},
  year={2024}
}

@inproceedings{zeng2024one,
  title={One-shot-but-not-degraded Federated Learning},
  author={Zeng, Hui and Xu, Minrui and Zhou, Tongqing and Wu, Xinyi and Kang, Jiawen and Cai, Zhiping and Niyato, Dusit},
  booktitle={Proceedings of the 32nd ACM International Conference on Multimedia},
  pages={11070--11079},
  year={2024}
}

@article{li2020practical,
  title={Practical one-shot federated learning for cross-silo setting},
  author={Li, Qinbin and He, Bingsheng and Song, Dawn},
  journal={arXiv preprint arXiv:2010.01017},
  year={2020}
}

@article{zhang2024one,
  title={One-shot federated learning via synthetic distiller-distillate communication},
  author={Zhang, Junyuan and Liu, Songhua and Wang, Xinchao},
  journal={Advances in Neural Information Processing Systems},
  volume={37},
  pages={102611--102633},
  year={2024}
}

@inproceedings{wanoasis,
  title={OASIS: One-Shot Federated Graph Learning via Wasserstein Assisted Knowledge Integration},
  author={Wan, Guancheng and Qian, Jiaru and Huang, Wenke and Xu, Qilin and Guo, Xianda and Li, Boheng and Zhang, Guibin and Du, Bo and Ye, Mang},
  booktitle={The Thirty-ninth Annual Conference on Neural Information Processing Systems},
year={2025}
}

@inproceedings{
qian2025ghost,
title={{GHOST}: Generalizable One-Shot Federated Graph Learning with Proxy-Based Topology Knowledge Retention},
author={Jiaru Qian and Guancheng Wan and Wenke Huang and Guibin Zhang and Yuxin Wu and Bo Du and Mang Ye},
booktitle={Forty-second International Conference on Machine Learning},
year={2025},
url={https://openreview.net/forum?id=nAk0ENu8LS}
}

@inproceedings{klicpera2019predict,
  title={Predict then Propagate: Graph Neural Networks meet Personalized PageRank},
  author={Klicpera, Johannes and Bojchevski, Aleksandar and G{\"u}nnemann, Stephan},
  booktitle={International Conference on Learning Representations (ICLR)},
  year={2019}
}

@inproceedings{mccallum2000automating,
  title={Automating the construction of internet portals with machine learning},
  author={McCallum, Andrew Kachites and Nigam, Kamal and Rennie, Jason and Seymore, Kristie},
  booktitle={Information Retrieval},
  volume={3},
  number={2},
  pages={127--163},
  year={2000},
  publisher={Springer}
}

@inproceedings{giles1998citeseer,
  title={CiteSeer: An automatic citation indexing system},
  author={Giles, C Lee and Bollacker, Kurt D and Lawrence, Steve},
  booktitle={Proceedings of the third ACM conference on Digital libraries},
  pages={89--98},
  year={1998}
}

@article{sen2008collective,
  title={Collective classification in network data},
  author={Sen, Prithviraj and Namata, Galileo and Bilgic, Mustafa and Getoor, Lise and Galligher, Brian and Eliassi-Rad, Tina},
  journal={AI magazine},
  volume={29},
  number={3},
  pages={93--93},
  year={2008}
}

@article{shchur2018pitfalls,
  author       = {Oleksandr Shchur and
                  Maximilian Mumme and
                  Aleksandar Bojchevski and
                  Stephan G{\"{u}}nnemann},
  title        = {Pitfalls of Graph Neural Network Evaluation},
  journal      = {CoRR},
  volume       = {abs/1811.05868},
  year         = {2018},
  url          = {http://arxiv.org/abs/1811.05868},
  eprinttype    = {arXiv},
  eprint       = {1811.05868},
  bibsource    = {dblp computer science bibliography, https://dblp.org}
}

@inproceedings{hu2020open,
  author       = {Weihua Hu and
                  Matthias Fey and
                  Marinka Zitnik and
                  Yuxiao Dong and
                  Hongyu Ren and
                  Bowen Liu and
                  Michele Catasta and
                  Jure Leskovec},
  editor       = {Hugo Larochelle and
                  Marc'Aurelio Ranzato and
                  Raia Hadsell and
                  Maria{-}Florina Balcan and
                  Hsuan{-}Tien Lin},
  title        = {Open Graph Benchmark: Datasets for Machine Learning on Graphs},
  booktitle    = {Advances in Neural Information Processing Systems 33: Annual Conference
                  on Neural Information Processing Systems 2020, NeurIPS 2020, December
                  6-12, 2020, virtual},
  year         = {2020},
  url          = {https://proceedings.neurips.cc/paper/2020/hash/fb60d411a5c5b72b2e7d3527cfc84fd0-Abstract.html},
  bibsource    = {dblp computer science bibliography, https://dblp.org}
}

@article{mernyei2020wikics,
  author       = {P{\'{e}}ter Mernyei and
                  Catalina Cangea},
  title        = {Wiki-CS: {A} Wikipedia-Based Benchmark for Graph Neural Networks},
  journal      = {CoRR},
  volume       = {abs/2007.02901},
  year         = {2020},
  url          = {https://arxiv.org/abs/2007.02901},
  eprinttype    = {arXiv},
  eprint       = {2007.02901},
  bibsource    = {dblp computer science bibliography, https://dblp.org}
}

@inproceedings{reddi2021adaptive,
  author       = {Sashank J. Reddi and
                  Zachary Charles and
                  Manzil Zaheer and
                  Zachary Garrett and
                  Keith Rush and
                  Jakub Kone{\v{c}}n{\'y} and
                  Sanjiv Kumar and
                  Hugh Brendan McMahan},
  title        = {Adaptive Federated Optimization},
  booktitle    = {9th International Conference on Learning Representations, {ICLR} 2021,
                  Virtual Event, Austria, May 3-7, 2021},
  publisher    = {OpenReview.net},
  year         = {2021},
  url          = {https://openreview.net/forum?id=LkFG3lB13U5},
  bibsource    = {dblp computer science bibliography, https://dblp.org}
}

@article{li2023fedgta,
  title={FedGTA: Topology-Aware Averaging for Federated Graph Learning},
  author={Li, Xunkai and Wu, Zhengyu and Zhang, Wentao and Zhu, Yinlin and Li, Rong-Hua and Wang, Guoren},
  journal={Proceedings of the VLDB Endowment},
  volume={17},
  number={1},
  pages={41--50},
  year={2023},
  publisher={VLDB Endowment}
}

@article{zhang2022dense,
  title={Dense: Data-free one-shot federated learning},
  author={Zhang, Jie and Chen, Chen and Li, Bo and Lyu, Lingjuan and Wu, Shuang and Ding, Shouhong and Shen, Chunhua and Wu, Chao},
  journal={Advances in Neural Information Processing Systems},
  volume={35},
  pages={21414--21428},
  year={2022}
}

@inproceedings{heinbaugh2023data,
  title={Data-free one-shot federated learning under very high statistical heterogeneity},
  author={Heinbaugh, Clare Elizabeth and Luz-Ricca, Emilio and Shao, Huajie},
  booktitle={The Eleventh International Conference on Learning Representations},
  year={2023}
}

\clearpage
% 插入附录 PDF 的所有页面，pages=- 表示从第一页到最后一页完整插入
\includepdf[pages=-]{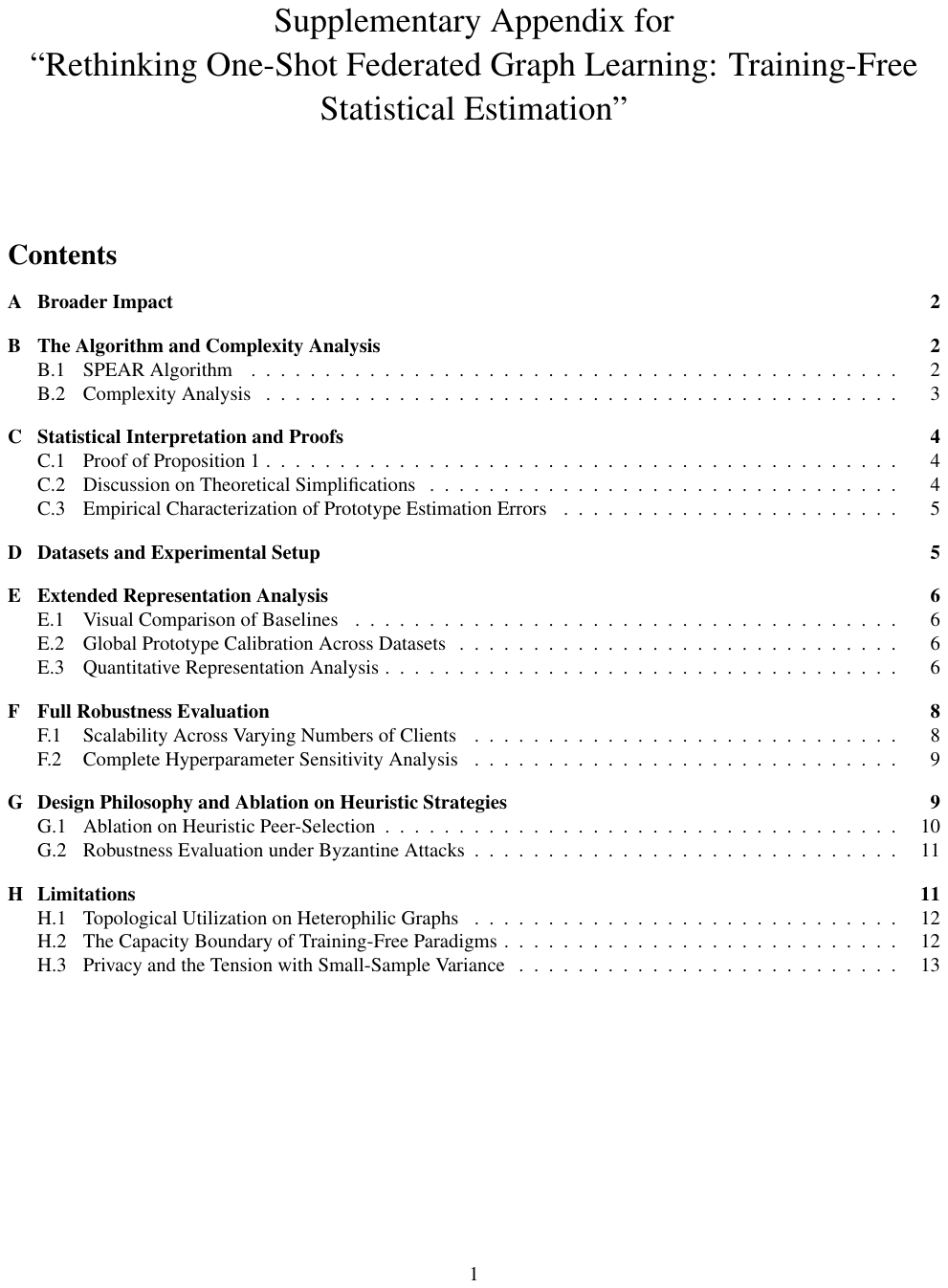}


\end{document}